\documentclass[letterpaper]{article}
\usepackage{proceed2e}
\usepackage[margin=1in]{geometry}

\usepackage{times}

\usepackage{graphicx} % more modern
\usepackage{subfigure} 

\usepackage{algorithm}
\usepackage{algorithmic}
\usepackage{amsmath}

\usepackage{tabularx}
\usepackage{multirow}

\usepackage{amsfonts}
\newcommand{\E}[1]{{\mathbb E}\left[ #1 \right]}
\usepackage{hyperref}

\title{Clustering-based Source-aware Assessment of \\True Robustness for Learning Models}

\author{ {\bf Ozsel Kilinc} \\
Electrical Engineering Dept. \\
University of South Florida\\
Tampa, FL 33620 \\
\And
{\bf Ismail Uysal}  \\
Electrical Engineering Dept.\\
University of South Florida \\
Tampa, FL 33620 \\
}

\begin{document}

\maketitle

\begin{abstract}
We introduce a novel validation framework to measure the true robustness of learning models for real-world applications by creating source-inclusive and source-exclusive partitions in a dataset via clustering.    We develop a robustness metric derived from source-aware lower and upper bounds of model accuracy even when data source labels are not readily available. We clearly demonstrate that even on a well-explored dataset like MNIST, challenging training scenarios can be constructed under the proposed assessment framework for two separate yet equally important applications: i) more rigorous learning model comparison and ii) dataset adequacy evaluation.  In addition, our findings not only promise a more complete identification of trade-offs between model complexity, accuracy and robustness but can also help researchers optimize their efforts in data collection by identifying the less robust and more challenging class labels. 
\end{abstract}

\section{INTRODUCTION}

The main goal in machine learning is to obtain a predictive model performing well on the input sample it has never seen before. This is called generalization and it separates machine learning from an optimization problem trying to find the minimum training error \cite{Goodfellow-et-al-2016}. Statistical learning theory \cite{vapnik91} defines this objective as risk minimization. However, expected risk cannot be directly calculated since exact data distribution is unknown and only available information is contained in a finite dataset, which is, in theory, an infinitesimally small sample of an otherwise infinite population. Instead, it is estimated by the empirical risk, i.e. the mean error over a test set whose examples are specifically not introduced during training. The assumption that among all the family of functions that can be modeled by the algorithm based on the training set, the one minimizing the empirical risk also minimizes the expected risk, turns the overall learning procedure into an empirical risk minimization problem. Empirical risk minimization also assumes that the examples in training and test sets are independent and identically distributed, i.e. i.i.d samples. Under these two assumptions, one can divide the dataset into training and test subsets and try to obtain the model providing the best generalization by validating model performance on the test subset. There exist a variety of partitioning methods commonly used in the literature most of which depend on random selection of samples in the dataset.

Holdout method is the simplest partitioning technique which divides the dataset into mutually exclusive training and test sets, however it may imply statistical uncertainty around the empirical risk, specifically when the chosen test set is small. Furthermore, since it does not provide any variance with respect to the training set, when comparing different algorithms, holdout method may not be considered optimal \cite{Dietterich98}. In such cases, computer intensive resampling methods can be used such as folded cross-validation, random subsampling or bootstrapping \cite{Kohavi95}.

In random subsampling, the holdout method is repeated $k$ times and the results of these runs are averaged. Unlike random subsampling, in bootstrapping, dataset is sampled with replacement, hence the resample size can be greater than the sample size. In cross-validation methods such as $k$-fold, the dataset is randomly divided into $k$ non-overlapping subsets, one of which is used as test set and the remaining $k-1$ subsets are used for the training. This procedure is repeated $k$ times to test the model on a different subset every time and the errors are averaged to estimate the overall test error \cite{Kohavi95}. It is shown that cross-validation provides an unbiased estimate of the test error but its variance may be very large \cite{Breiman96}. Also, \cite{BengioG04} studied the problem of that there exists no universal unbiased estimator of the variance of $k$-fold cross-validation.

Despite the claim of learning theory that a machine learning algorithm can generalize reasonably well using a finite training set, the ``no free lunch'' theorem \cite{wolpert1996lack} states that when averaged over all possible data-generating distributions, every classification algorithm has the same generalization potential for the previously unobserved points. However, the motivation behind designing learning algorithms is to make the best assumptions about the kinds of probability distributions we might encounter in real-world applications. In other words, the goal of machine learning is not to seek a universally optimal learning algorithm, but to understand the relevant real-world data distributions and design algorithms performing well on data drawn from them \cite{Goodfellow-et-al-2016}.

Even under these assumptions, we cannot ignore the possibility that the AI agent we have designed might encounter data drawn from a different kind of data-generating distribution which may not have been available in our finite dataset used for training. Hence, there is need for evaluating the designed algorithms under such worst-case scenarios especially for critically deployed applications. This should, of course, be relevant to the use-case the algorithm is designed for, i.e. a hand-written digit classifier should not be tested with a picture of an animal, instead its performance should be measured on a different hand-written digit dataset whose examples are drawn from a distribution not used for training and testing. Collecting a secondary dataset for this purpose is not feasible or practical for every single learning problem. A more intelligent solution to test for these worse-case scenarios would be to find the examples belonging to different distributions in the original dataset and make the dataset partitioning according to this information. The assumption in this case is that since the model is trained and tested using less likely examples, obtained generalization will be a better representation of worst-case scenarios. 

\cite{KilincU15} studied this problem on a small activity recognition dataset where user IDs of the volunteers performing the activities were available for each sample along with the activity labels. Assuming that each user, i.e. data source, generates different distributions for the same activity label, they proposed two types of source-aware partitioning. While \textit{inclusive source-aware} ensures that samples from each source, i.e. data-generating process, are arranged to be included in all subsets, \textit{exclusive source-aware} ensures mutually exclusive use of these sources. Validated results through different resampling methods show that while \textit{inclusive source-aware} type partitioning ends up with the estimation of test error close to the one obtained without using source-awareness, i.e. completely randomized division of training and test subsets, \textit{exclusive source-aware} partitioning results in significantly higher error rates, which may possibly be interpreted as a better estimation of the model generalization when it encounters samples less similar to those in the existing dataset.

Unfortunately, not all datasets have this kind of labeled information about the source identities. In this paper, we propose to use a clustering algorithm to artificially find different sources within the dataset to generalize the use of source-aware partitioning to all datasets. Moreover, in order to obtain a more robust evaluation of different learning models, we propose the use of these two estimations of expected generalization, i.e. test accuracy obtained using \textit{inclusive source-aware} and \textit{exclusive source-aware} data partitioning which respectively simulate the best-case and the worst-case scenarios that might be encountered in real-world applications. We also introduce a similar comparison to be made on the actual adequacy of the dataset itself to see if it's sufficient to obtain a good enough generalization or if more samples need to be collected.  We used MNIST \cite{lecun1998mnist} to show that dataset partitioning that takes the source distribution into account is not only crucial for small datasets, but even for larger, and well-explored datasets such as MNIST  where state-of-the-art solutions can readily obtain error rates around 0.21\%. 

This paper is organized as follows. In the following section, we explain the clustering methods applied to find source distributions. The third section describes how the obtained clusterings are used to implement source-aware partitioning. Then, the fourth section introduces the proposed robustness metrics and framework. In the fifth section, results obtained for model robustness and dataset adequacy are presented and then, the paper is concluded with final comments.

\section{USING CLUSTERING TO FIND SOURCE DISTRIBUTIONS}

Any clustering algorithm can be applied to find the different sources, i.e. data generating processes or distributions within the dataset. Samples of each class are clustered into $k$ clusters where each cluster is assumed to correspond to a different source distribution within the same class label. The hypothesis is that, for each class, the samples in different clusters are generated by different data generating processes and if the samples of these clusters are completely isolated from the training set, yet included in the test set, the generalization of the model on these samples will get worse.

\begin{figure}[t]
	\begin{center}
		\centerline{\includegraphics[width=\columnwidth,trim={0.2cm 0.2cm 0.2cm 0.2cm},clip]{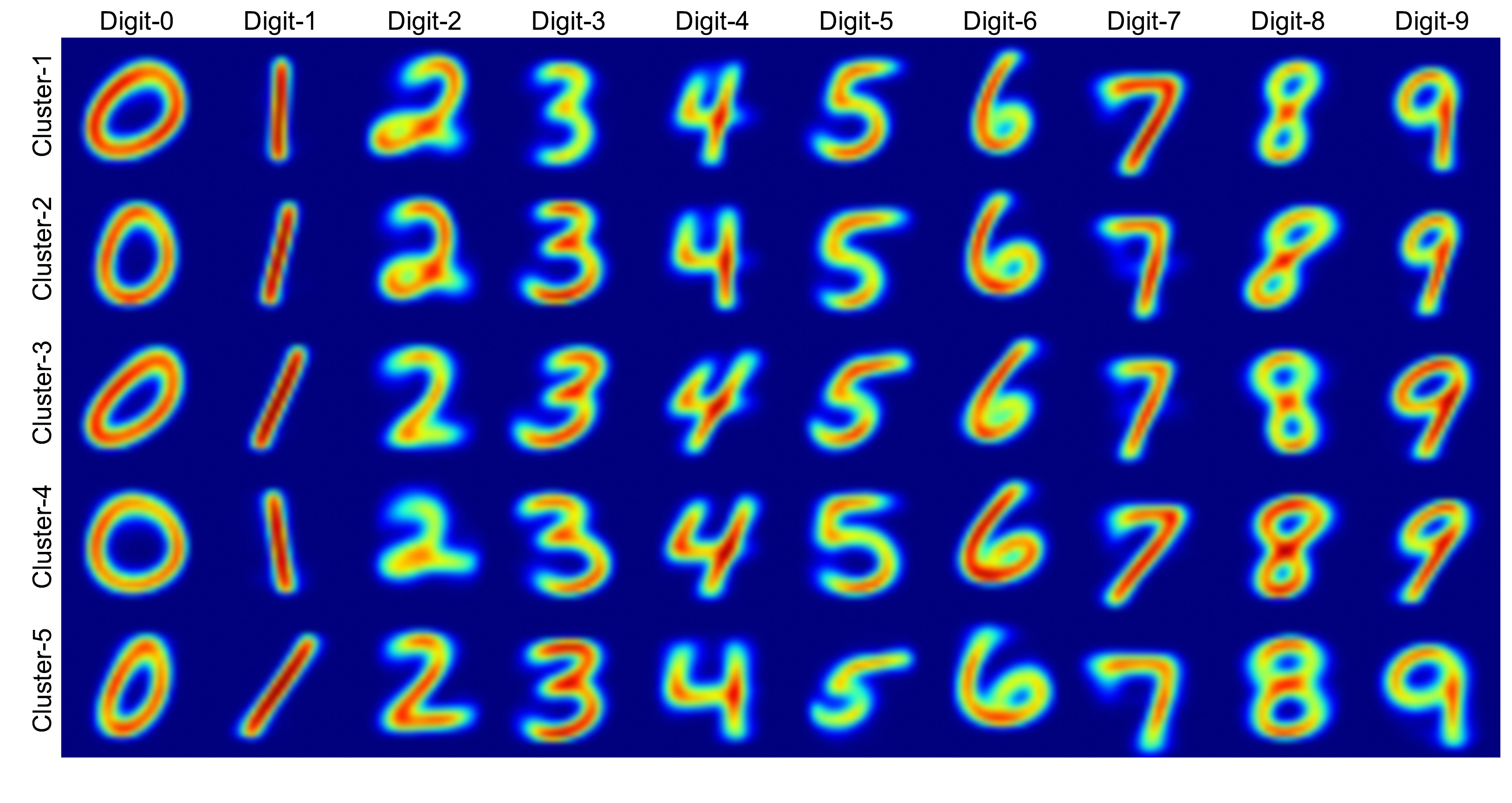}}
		\caption{Visualization of the means of clusters found by using $k$-means clustering algorithm where $k=5$ }
		\label{fig:clusters_kmeans}
	\end{center}
\end{figure}

We cluster the each one of the 10 digit classes of MNIST dataset using $k$-means clustering algorithm \cite{kmeans} by choosing $k=5$ to divide the dataset into approximately 20\% sized clusters. Since the training and test sets will be reconstructed based on source-aware partitioning, we abandon MNIST's existing partitioning to combine existing training and test sets and apply partitioning to all of the 70,000 available samples. Figure~\ref{fig:clusters_kmeans} visualizes the means of clusters found by using $k$-means.  

MNIST dataset happens to have some good features which help $k$-means obtain good clusters. Namely, the images are centered on a 28x28 plane with respect to the center of mass of the pixels. Besides, since we are interested in clustering among the samples of each digit where variance between samples is significantly small, this creates a considerably simple clustering problem which can easily be handled by the relatively simple computation of $k$-means. However, $k$-means may not always perform optimally when doing in-class clustering for more complex datasets. 

\cite{KilincU17deepclustering} recently proposed a method called \textit{auto-clustering output layer} (ACOL) which enables deep neural networks to find the subclasses of parent classes. ACOL can be used for semi-supervised problems where labeling is partially available or for completely unsupervised problems using the proposed ``pseudo-class'' trick. It is shown that ACOL outperforms $k$-means by a significantly large margin in clustering applications created on MNIST for both types of learning. They state that ACOL enables the use of deep learning models - originally proposed for supervised classification - for clustering problems and claim that this paradigm shift can be called ``deep clustering" because the capacities of resulting clustering models can be increased by adding additional layers as we do for deep learning classifiers. 

\begin{figure}[t]
	\begin{center}
		\centerline{\includegraphics[width=\columnwidth,trim={0.2cm 0.2cm 0.2cm 0.2cm},clip]{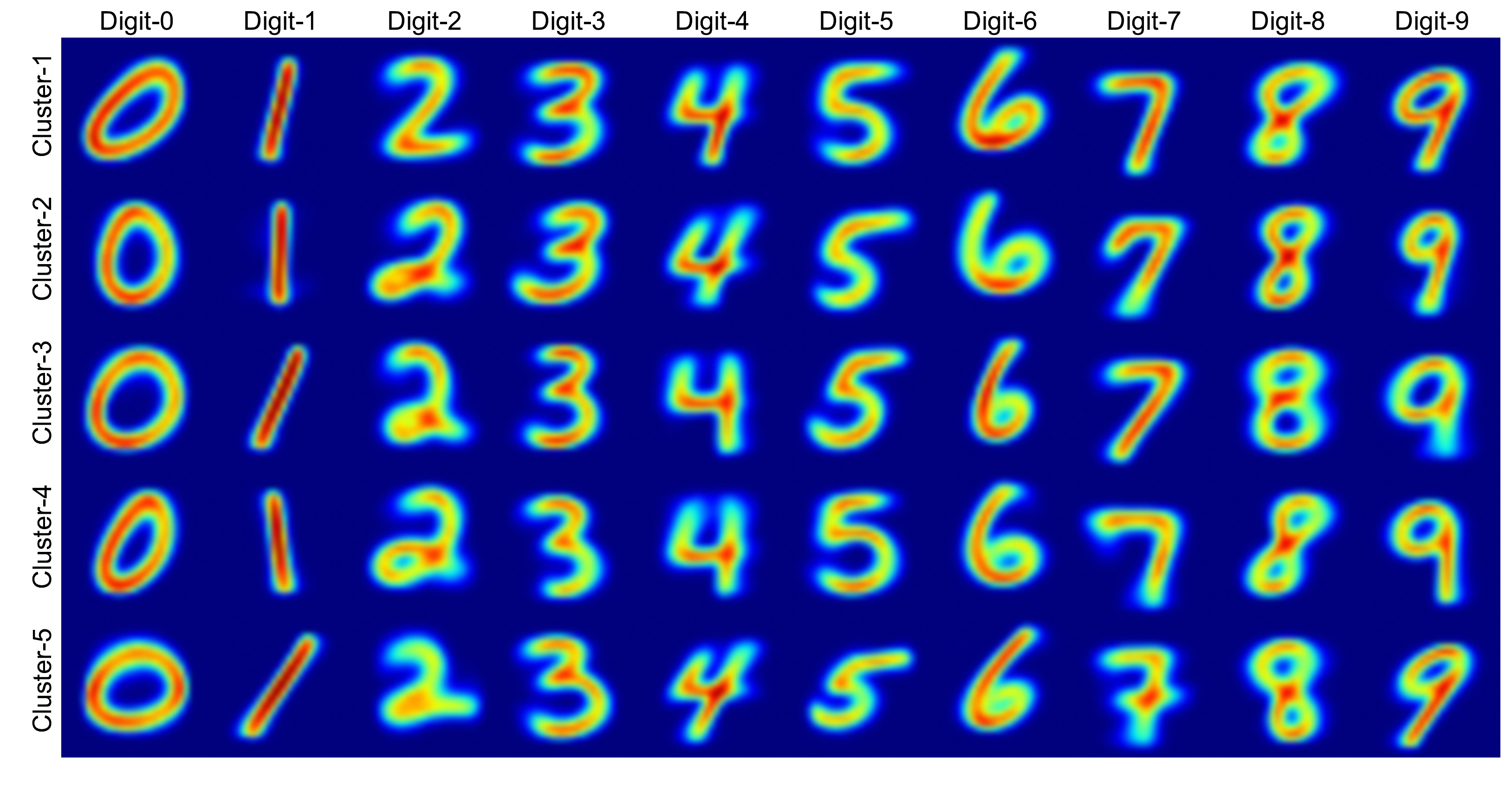}}
		\caption{Visualization of the means of clusters found by using ACOL}
		\label{fig:clusters}
	\end{center}
\end{figure}

To demonstrate that the effects of source-aware partitioning are independent of the clustering algorithm used, we have also applied ACOL with $K=5$ and other hyperparameters as described in \cite{KilincU17deepclustering} to obtain in-class clustering. Clustering is performed on a convolutional network model we labeled ``CNN-2'' whose specifications are described in the subsequent sections along with other models (such as support vector machines, or other neural network architectures) used in experiments. Similar to Figure~\ref{fig:clusters_kmeans}, Figure~\ref{fig:clusters} shows the means of clusters found by using ACOL clustering. We have used the silhouette score \cite{Rousseeuw87silhouettes} to evaluate the consistency within formed clusters in which $k$-means and ACOL clustering get the scores of 0.0456 and 0.0280 respectively. Hence, we may possibly observe more significant differences between \textit{inclusive source-aware} and \textit{exclusive source-aware} partitioning when $k$-means generated clustering is used. It is important to point out that traditional clustering performance cannot be computed in this case since ground-truth labels are simply not known. The main goal is to create more challenging subsets than simply using random resampling or hold-out as MNIST can already be considered as having a challenging test set because sets of writers for its training set and test set are disjoint. Clustering in this case can further help us find the groups of writers generating similar digits (in terms of roundness, tilt, etc.) and apply partitioning based on this information.

In the end, both schemes are used to apply source-aware partitioning. The following section describes how we apply \textit{inclusive} and \textit{exclusive} source-awareness using these clusters. 

\section{APPLYING SOURCE-AWARE PARTITIONING}

In source-aware partitioning, assuming that each one of the 5 clusters obtained via k-means or ACOL correspond to a different source which generates data with a different distribution, training and test subsets are constructed based on this information instead of completely randomized sampling. 

More specifically, for \textit{exclusive source-aware} partitioning, for each one of the 10 digit classes, one of the 5 clusters is randomly chosen and all its samples are added to the test set while the samples of the remaining 4 clusters are added to training set effectively creating a 20\% to 80\% split. This partitioning ensures that the samples of the selected test cluster for each class are never introduced to the algorithm during training. Random selection of the clusters added to test set is performed independently for each class. Since the correlations between the clusters of different classes are not known, i.e. we do not know that samples in which cluster of digit-1 are generated by the same source generating the samples in the first cluster of digit-0, we cannot cover all combinations in 5 trials as we do in $k$-fold cross-validation. Instead, it requires $5^{10}$ combinations for 5 clusters and 10 class labels, which is computationally almost impossible to cover completely. Hence, we randomly picked clusters for each class and repeated the same experiment 100 times to satisfy the desired statistical significance.

For \textit{inclusive source-aware} partitioning, 20\% of the samples from each of the 5 clusters are randomly chosen and added to test set while the remaining 80\% of samples are added to training set. Although the resulting subsets are also mutually exclusive, i.e. a sample is added either to training set or test set – but not both at the same time, in terms of source distributions, they are forced to have the same or similar distributions. In other words, this partitioning ensures that both training and test set involve samples from all source distributions. Similar to \textit{exclusive source-aware} random selection of the clusters added to test set is performed independently for each class instead of $k$-fold cross-validation due to the number of different combinations possible.  Each experiment for \textit{inclusive source-aware} partitioning has also been repeated for 100 times as we do for \textit{exclusive source-aware} partitioning for desired statistical significance.

\section{PROPOSED ROBUSTNESS METRICS AND FRAMEWORK}

We claim that the accuracy obtained using \textit{exclusive source-aware} partitioning, $Acc_X$, is a better estimation for the expected accuracy of the trained algorithm when it encounters unlikely examples compared to those we have used for training, i.e. worst-case scenario, and in the same manner, accuracy obtained using \textit{inclusive source-aware} partitioning, $Acc_I$, is a better estimation for the expected accuracy in the best-case scenario. Hence, we propose to use an interval, instead of a single value, to define the expected real-world performances of the learning algorithms such that:

\begin{equation}
\textit{Accuracy Interval} = \mathcal{I} := \big[Acc_X, Acc_I\big]
\end{equation}

Since we observe natural variance in obtained accuracies with respect to sampled training and test sets for the 100 repeated trials, it is more accurate to define this interval as \textit{expected accuracy interval} such that

\begin{equation}
\bar{\mathcal{I}} = {\mathbb E}[\text{\boldmath$\mathcal{I}$}] = \big[\E{\textbf{\textit{Acc}}_\textbf{\textit{X}}}, \E{\textbf{\textit{Acc}}_\textbf{\textit{I}}}\big]
\end{equation}

where we expect to observe $\E{\textbf{\textit{Acc}}_\textbf{\textit{I}}} \ge \E{\textbf{\textit{Acc}}_\textbf{\textit{X}}}$ given that the source identification process, i.e. applied clustering algorithm, is successful in distinguishing the sources. We also expect to observe that $\E{\textbf{\textit{Acc}}_\textbf{\textit{I}}}$ is approximately equal to the expected accuracy obtained using a completely random partitioning method, $\E{\textbf{\textit{Acc}}}$, i.e. without using any source awareness, if the dataset is sufficiently large. Otherwise, $\E{\textbf{\textit{Acc}}_\textbf{\textit{I}}} > \E{\textbf{\textit{Acc}}}$
since there might possibly exist cases where samples of a source are mostly isolated from the training set during source-unaware random sampling.

In order to compare the learning algorithms with respect to stability in their expected real-life performances, i.e. robustness, we can use the range of the \textit{expected accuracy interval} and normalize it by $\E{\textbf{\textit{Acc}}_\textbf{\textit{I}}}$ to obtain a metric scaled between 0 and 1. Hence, \textit{robustness} becomes 
\begin{equation}
\rho := 
1-\frac{\E{\textbf{\textit{Acc}}_\textbf{\textit{I}}}-\E{\textbf{\textit{Acc}}_\textbf{\textit{X}}}}{\E{\textbf{\textit{Acc}}_\textbf{\textit{I}}}} = 
\frac{\E{\textbf{\textit{Acc}}_\textbf{\textit{X}}}}{\E{\textbf{\textit{Acc}}_\textbf{\textit{I}}}}
\end{equation}

which emphasis the narrowness of the \textit{expected accuracy interval} regardless of the magnitudes of the obtained accuracies.

\section{RESULTS}

In this section we will present detailed analysis of the proposed partitioning algorithms for two separate applications: i) measuring learning model robustness and ii) measuring dataset robustness.

\subsection{MODEL ROBUSTNESS}

We compare the following learning models.

\begin{small}
\begin{itemize}
	\item \textbf{SVM:}
	\begin{itemize}
		\item C = 1, kernel = rbf, gamma = 0.01
	\end{itemize}		
\item \textbf{MLP:} 
	\begin{itemize}
	\item Feedforward 2048 - 50\% Dropout - maxnorm(2)
	\item Feedforward 2048 - 50\% Dropout - maxnorm(2)
	\item Feedforward 2048 - 50\% Dropout - maxnorm(2)
	\end{itemize}
\item \textbf{CNN-1:}
	\begin{itemize}
	 \item 32x3x3 - 32x3x3 - MP2x2 - 25\% Dropout 
	 \item Feedforward 2048 - 50\% Dropout
	\end{itemize}
\item \textbf{CNN-2:}
	\begin{itemize}
	\item 32x3x3 - 32x3x3 - MP2x2 - 25\% Dropout 
	\item 64x3x3 - 64x3x3 - MP2x2 - 25\% Dropout 
	\item Feedforward 2048 - 50\% Dropout
	\end{itemize}
\item \textbf{CNN-3:}
\begin{itemize}
	\item 32x3x3 - 32x3x3 - MP2x2 - 25\% Dropout 
	\item 64x3x3 - 64x3x3 - MP2x2 - 25\% Dropout 
	\item 128x3x3 - 128x3x3 - MP2x2 - 25\% Dropout 
	\item Feedforward 2048 - 50\% Dropout
\end{itemize}	
\end{itemize}
\end{small}

For all the experiments, we have trained all the neural network models for the same number of epochs which is sufficiently large to ensure that testing error stops decreasing for all models in all cases. To eliminate the effect of overtraining, especially in smaller models, we have recorded the maximum testing accuracy observed during each training. Evaluating 5 models (SVM, MLP, CNN-1, CNN-2 and CNN-3) using 2 types of partitioning (\textit{inclusive source-aware} and \textit{exclusive source-aware} and 2 different kinds of cluster (k-means and ACOL) result in 5 x 2 x 2 = 20 sets of experiments which are then repeated 100 times for desired statistical significance. 

\begin{figure}[ht]
	\begin{center}
		\centerline{\includegraphics[width=\columnwidth,trim={0cm 0cm 0cm 0cm},clip]{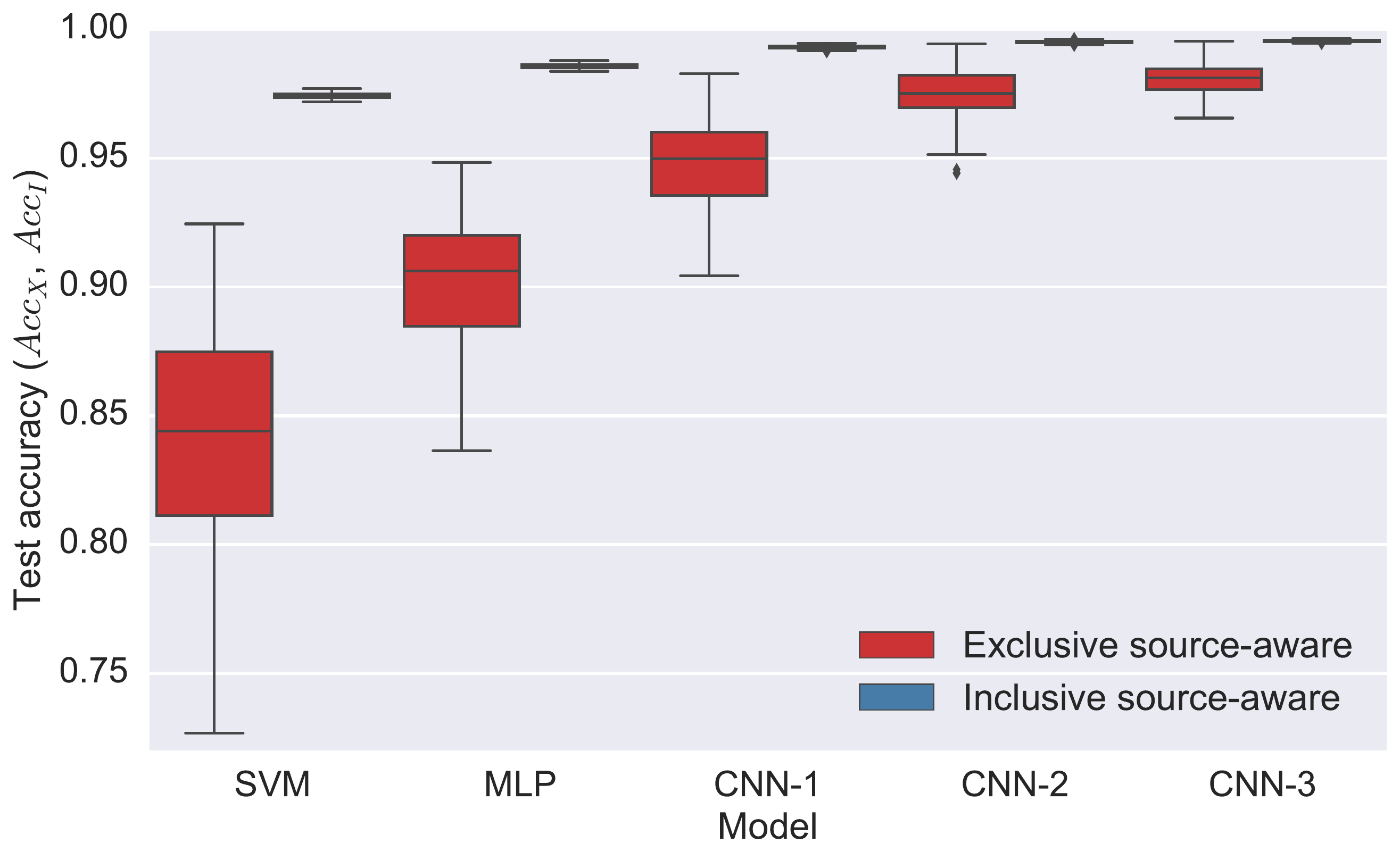}}
		\caption{Box plot of test accuracies observed in experiments where source-aware partitioning is performed according to the clustering scheme obtained by ACOL.}
		\label{fig:model_acc_box}
	\end{center}
	\vskip -0.2in
\end{figure}

Figure~\ref{fig:model_acc_box} presents the box plots of test accuracies where source-aware partitioning is performed according to the clustering scheme obtained using ACOL.  The figure very clearly shows that the test accuracies obtained using \textit{inclusive source-aware} partitioning, $Acc_I$, have a narrow dispersion, whereas, \textit{exclusive source-aware} partitioning results in a significantly wider distribution of test accuracies, $Acc_X$, due to the diversity of combinations observed when distributing the sources exclusively into training and test sets. Besides, while the expected value and variance of $Acc_I$ only slightly vary with respect to the learning algorithm, both expected value and variance of $Acc_X$ drastically improve with increasing model capacity to represent a better decision metric in determining which model outperforms the rest.

\begin{figure}[ht]
	\begin{center}
		\vskip 0.1in
		\centerline{\includegraphics[width=\columnwidth,trim={0.2cm 0.2cm 0.2cm 0.2cm},clip]{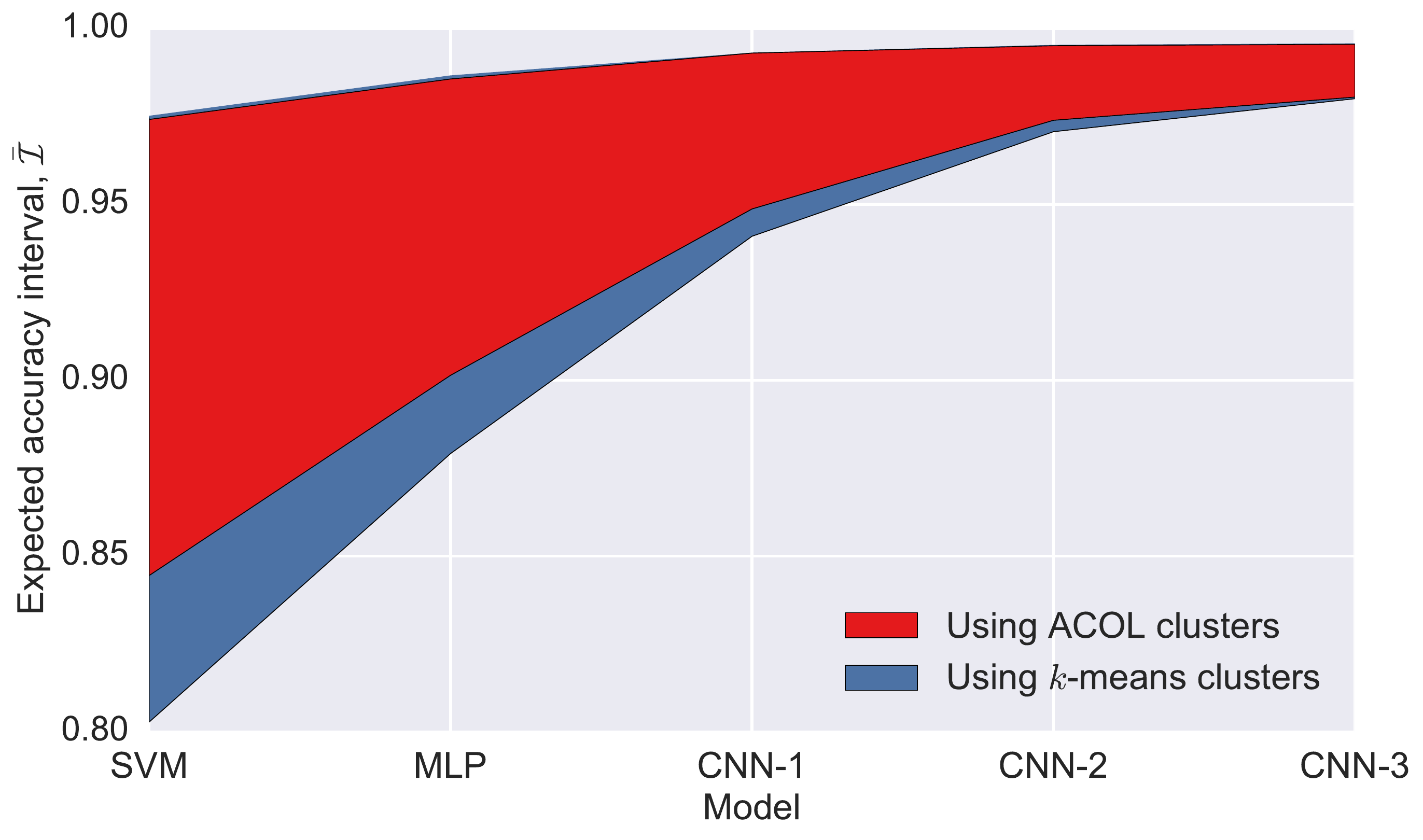}}
		\caption{Change of \textit{expected accuracy interval} with respect to the applied learning model type. The upper and lower bounds of the colored regions represent the expected $Acc_I$ and $Acc_X$ respectively.}
		\label{fig:comparison_interval}
	\end{center}
	\vskip -0.3in
\end{figure}

For better understanding, Figure~\ref{fig:comparison_interval} provides the visualization of \textit{expected accuracy interval} $\bar{\mathcal{I}}$, where the upper and lower bounds of the colored regions represent the expected $Acc_I$ and the expected $Acc_X$ respectively. We propose that these colored regions show the range of accuracies expected to be observed in real-world applications depending on the likelihood of the encountered inputs being similar to those included in our finite dataset. This figure illustrates two different intervals obtained using two different clustering schemes where red and blue colored regions represent clustering schemes obtained using ACOL and $k$-means respectively. Upper bounds of these two regions correlate well as both clustering schemes result in approximately equal expected $Acc_I$. The clustering obtained by $k$-means infer a wider interval at the bottom by yielding a lower $Acc_X$. One may interpret this difference as an indicator pointing that $k$-means provides a more challenging, yet not necessarily better, mutually exclusive separation of the sources during in-class clustering. Another observation is that the differences between lower bounds, together with the ranges of both regions, shrink as the model capacity increases.   

\begin{figure}[t]
	\begin{center}
		\centerline{\includegraphics[width=\columnwidth,trim={0.2cm 0.2cm 0.2cm 0.2cm},clip]{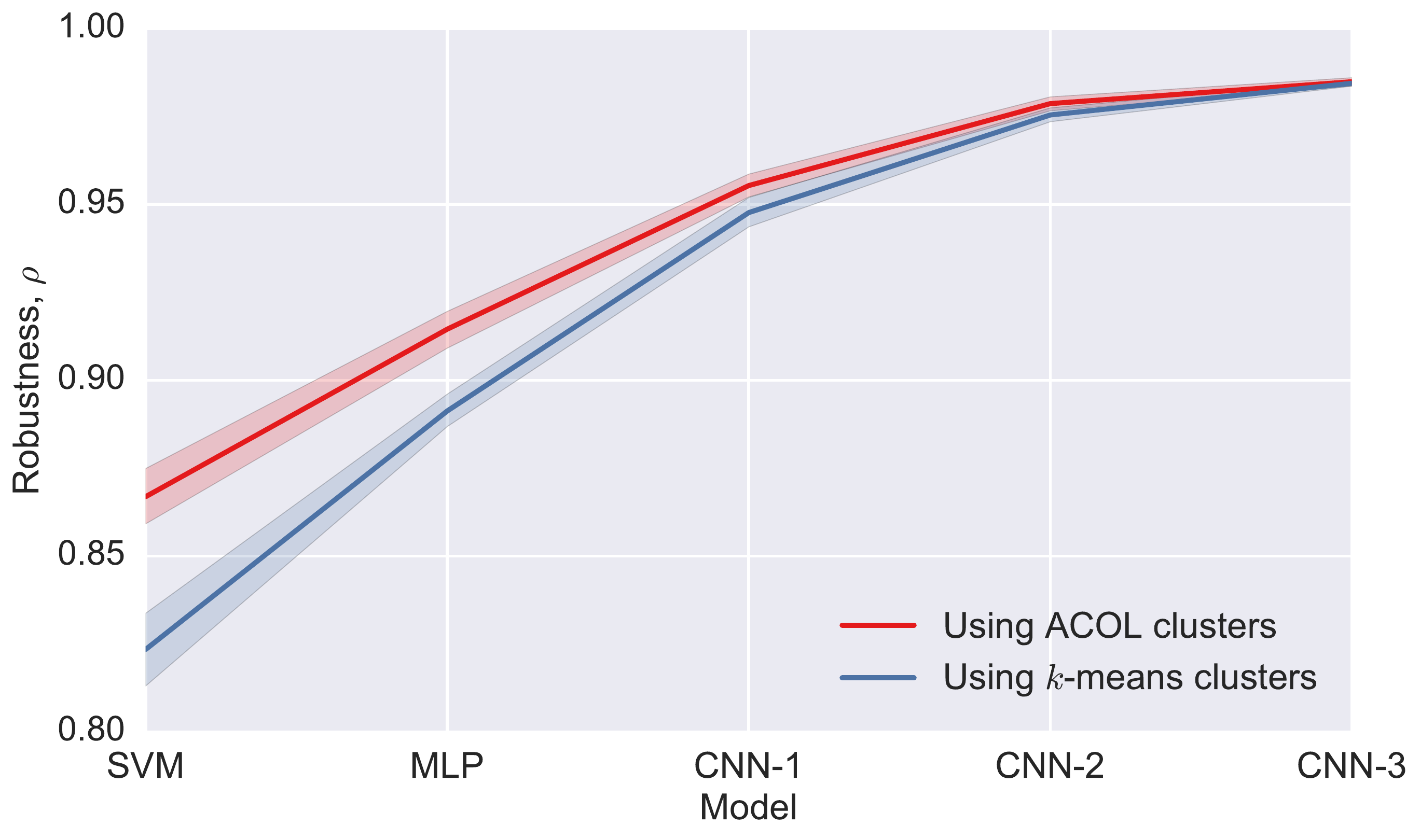}}
		\caption{Behavior of the proposed \textit{robustness} metric with respect to the applied learning model type.}
		\label{fig:comparison}
	\end{center}
	\vskip -0.2in
\end{figure}

Devised \textit{robustness} metric can be seen in Figure~\ref{fig:comparison} and Table~\ref{tab:model_sum} summarizes the results obtained on all models using both clustering approaches. Even as SVM and MLP algorithms obtain over 97\% accuracy when they are tested with likely examples, their generalizations on unlikely ones are clearly not as successful as CNN models, which is slightly more apparent when $k$-means clustering is used. Through their convolution layers, CNN models obtain a better representation of the visual input at the classifier layer, which might help extracting additional information to compensate for the effect of excluded sources. In fact, as seen in the table, each added convolution layer improves their generalization ability to unlikely examples, which evidently makes them more robust. Moreover, CNNs are also indifferent to the two clustering schemes whose effects are more apparent on \textit{exclusive source-aware} performances of SVM and MLP.

\begin{table}[ht]
	\begin{center}
		\caption{Summary of \textit{expected accuracy interval} and \textit{robustness} observed on all five models using both clustering schemes.}
		\label{tab:model_sum}
		\resizebox{\columnwidth}{!} {
			\begin{tabular}{ c|cc|cc }
				%\hline
				\multirow{2}{*}{\textbf{Model}} & \multicolumn{2}{c|}{\textbf{Using ACOL Clusters}} & \multicolumn{2}{c}{\textbf{Using $k$-mean Clusters}} \\ \cline{2-5} 
								& 							& 		  & 						&	\\[-0.9em] 				
								& $\bar{\mathcal{I}}$ (\%)  & $\rho$  & $\bar{\mathcal{I}}$ (\%)& $\rho$\\ \hline%\hline
								& 							& 		  & 						&	    \\[-0.9em] 
				\textbf{SVM} 	&	$\big[84.44,97.42\big]$ & $0.867$ &	$\big[80.28,97.50\big]$ & $0.823$\\ %\hline
								& 							& 		  & 						&	    \\[-0.9em] 
				\textbf{MLP} 	& 	$\big[90.14,98.57\big]$ & $0.915$ &	$\big[87.92,98.64\big]$ & $0.891$\\ %\hline
								& 							& 		  & 						&	    \\[-0.9em] 
				\textbf{CNN-1} 	& 	$\big[94.87,99.30\big]$ & $0.955$ &	$\big[94.09,99.29\big]$ & $0.948$\\ %\hline
								& 							& 		  & 						&	    \\[-0.9em] 
				\textbf{CNN-2} 	& 	$\big[97.39,99.51\big]$ & $0.979$ &	$\big[97.07,99.51\big]$ & $0.975$\\ %\hline
								& 							& 		  & 						&	    \\[-0.9em] 
				\textbf{CNN-3} 	& 	$\big[98.05,99.55\big]$ & $0.985$ &	$\big[98.01,99.55\big]$ & $0.984$\\ %\hline
				
			\end{tabular}
		}
	\end{center}
\end{table}

\subsection{DATASET ADEQUACY}

Using source-aware partitioning, one can also analyze whether the number of examples in the training dataset is enough to obtain good generalization on unlikely examples. This kind of analysis may be useful to help the researcher determine the necessity to collect more samples. To observe the effect of number of training samples, we reduced the size of the training set while keeping test set the same as in the experiments with the full sized training set. For these experiments, we fixed our learning models to CNN and applied partitioning using the ACOL clustering scheme. 

\begin{figure}[t]
	\begin{center}
		\centerline{\includegraphics[width=\columnwidth,trim={0cm 0cm 0cm 0cm},clip]{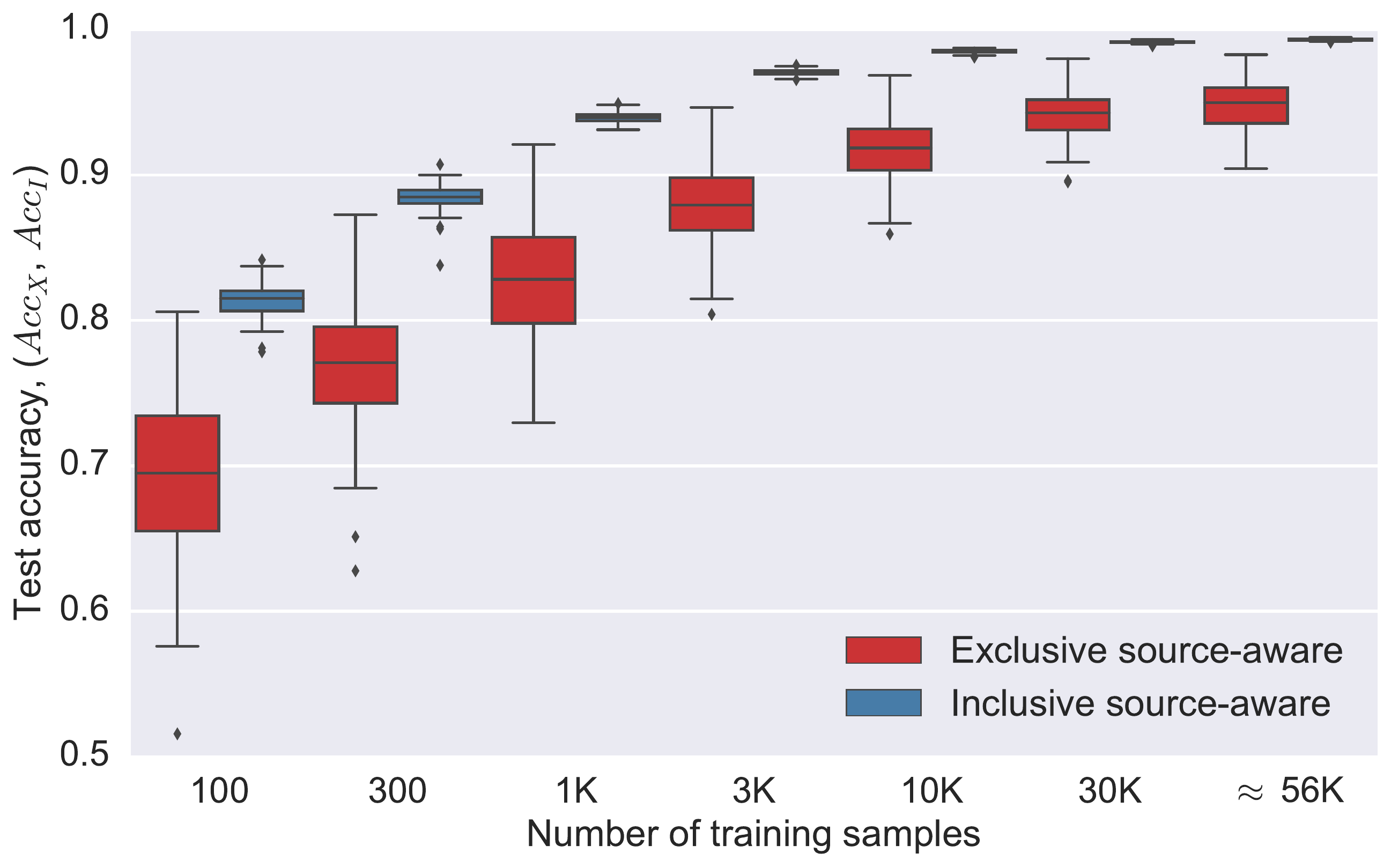}}
		\caption{Box plot of test accuracies observed on CNN-1 model in the experiments performed with varying training set sizes.}
		\label{fig:data_acc_box}
	\end{center}
	\vskip -0.2in
\end{figure}

\begin{figure}[t]
	\begin{center}
		\centerline{\includegraphics[width=\columnwidth,trim={0cm 0cm 0cm 0cm},clip]{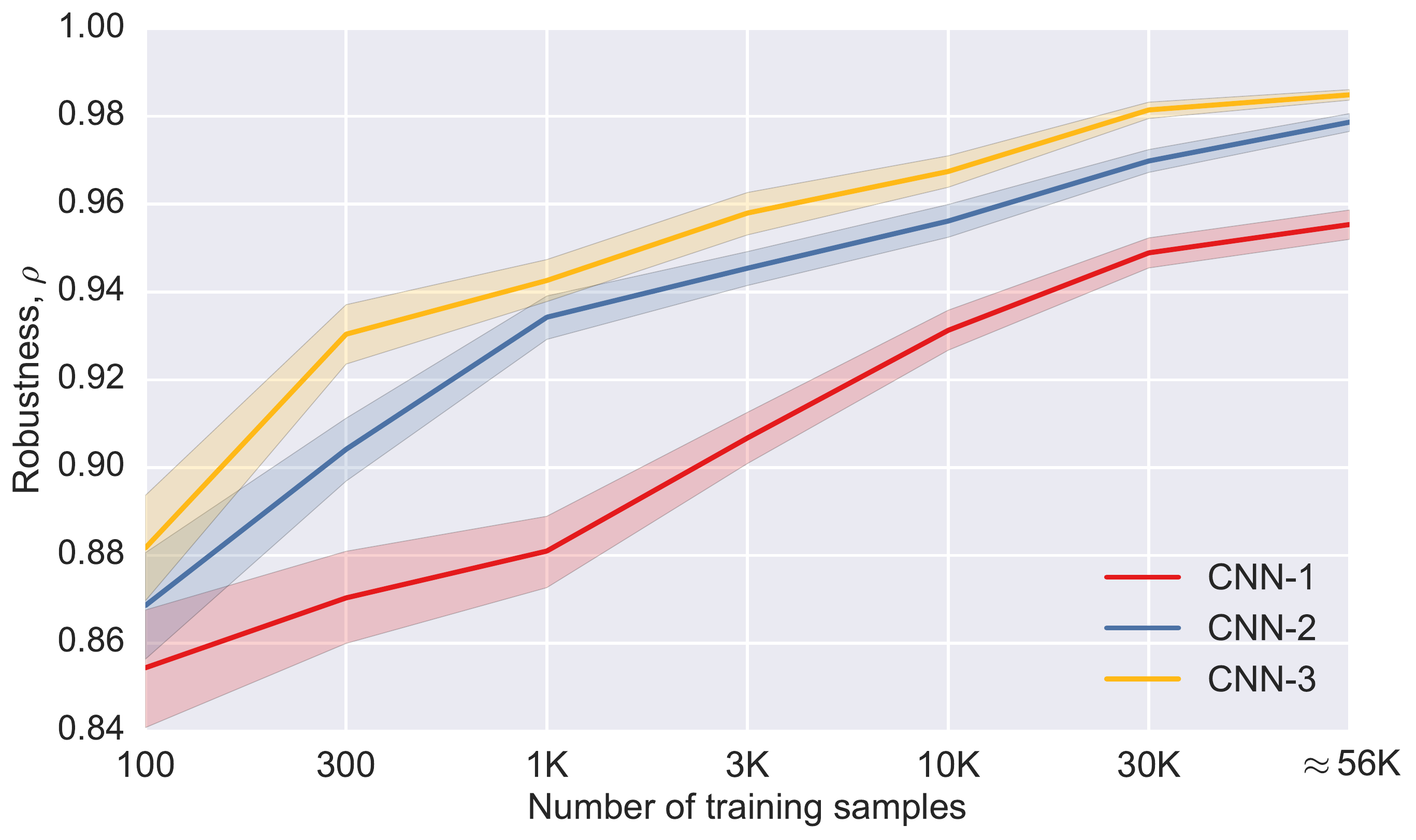}}
		\caption{Change of \textit{robustness} with respect to the number of training samples observed on all three CNN models.}
		\label{fig:data_robust}
	\end{center}
	\vskip -0.3in
\end{figure}

As an example, for CNN-1 model, box plots of test accuracies obtained in these experiments are given in Figure ~\ref{fig:data_acc_box}. One can observe that, for both $Acc_I$ and $Acc_X$, expected values increase and variance decrease with increasing training set size.
\begin{figure}[b]
	\begin{center}
		\centerline{\includegraphics[width=\columnwidth,trim={0cm 0cm 0cm 0cm},clip]{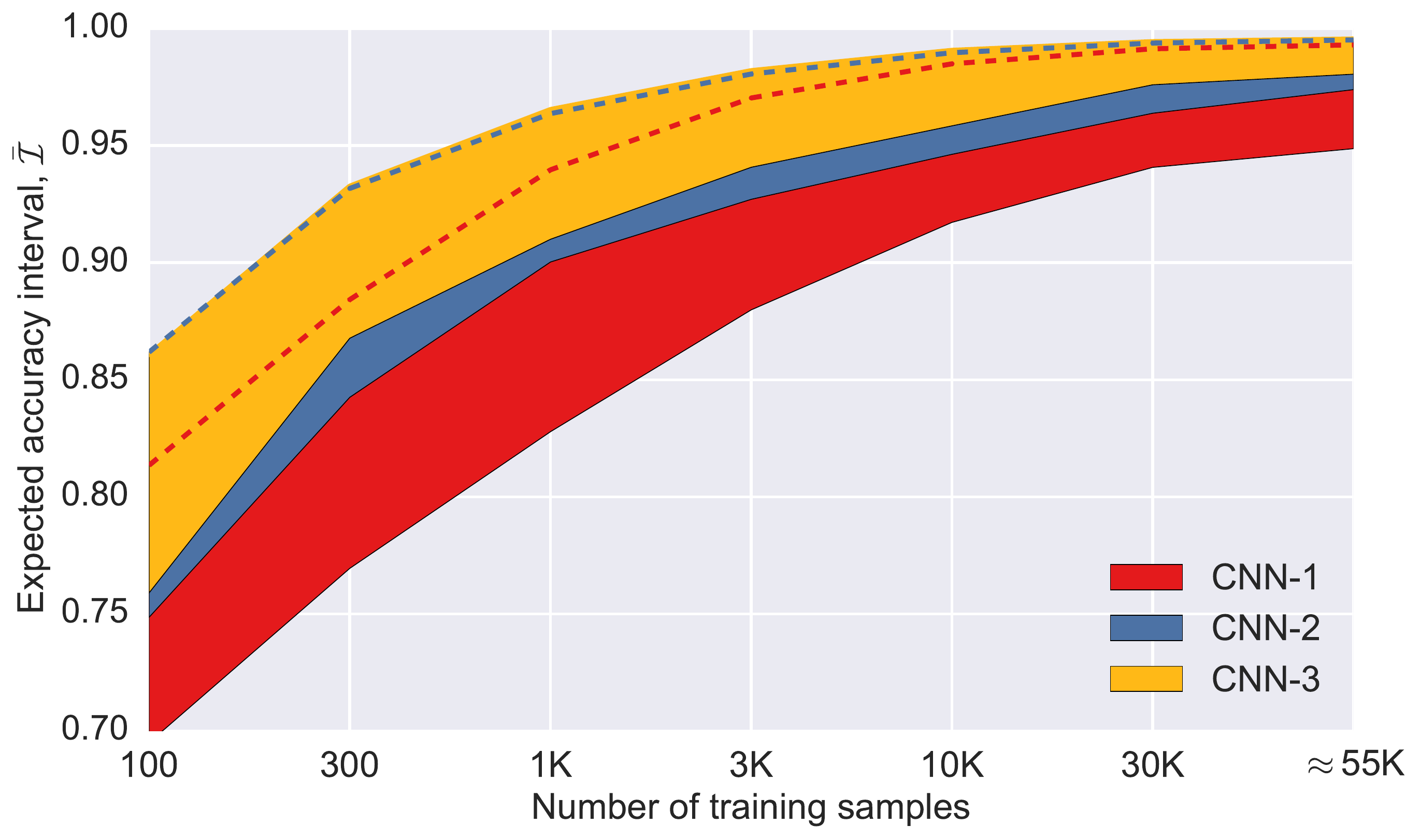}}
		\caption{Change of \textit{expected accuracy interval} with respect to the number of training samples observed on all three CNN models.}
		\label{fig:data_acc}
	\end{center}
	\vskip -0.1in
\end{figure} 
In order to interpret this behavior with respect to the model capacity, Figure~\ref{fig:data_acc} and Figure~\ref{fig:data_robust} respectively illustrate \textit{expected accuracy interval} and \textit{robustness} observed on all three CNN models. One can observe that to obtain a good expected generalization on likely examples, which is evaluated by $Acc_I$, using 10000 training samples is sufficient. However, further increases on the training set size improve the expected generalization on unlikely test examples, which is evaluated by $Acc_X$, and thus increases robustness. Another remarkable observation is that adding an additional convolution layer on CNN-2 has no effect on its generalization for likely examples (as shown by the blue dotted line coinciding with the boundary of CNN-3) but improves generalization for the unlikely samples. Considering that we expect to obtain $Acc \approx Acc_I$, without source-aware partitioning one observes almost no difference between using CNN-2 and CNN-3 and might consequently think that there is no need to use the bigger model. However, \textit{exclusive source-aware} performances of these two models clearly show CNN-3 outperforming CNN-2 through better generalization on unlikely samples as a better representation of expected real life performance. Table~\ref{tab:dataset_interval} and Table~\ref{tab:dataset_robustness} summarize the \textit{expected accuracy interval} and \textit{robustness} metrics observed on all three models respectively.

\begin{table}[ht]
\begin{center}
\caption{Summary of \textit{expected accuracy interval} observed on all three CNN models with varying training set size.}
\label{tab:dataset_interval}
\resizebox{\columnwidth}{!} {
\begin{tabular}{c|ccc}
\multirow{2}{*}{\textbf{\begin{tabular}[c]{@{}c@{}}\# of training \\samples\end{tabular}}} & \multicolumn{3}{c}{$\bar{\mathcal{I}}$ (\%)}                   \\ \cline{2-4} 
\\[-1em]
& CNN-1 & CNN-2 & CNN-3 \\ \hline
\\[-0.9em]
\textbf{100} & $\big[69.50,81.34\big]$ & $\big[74.85,86.18\big]$ & $\big[75.89,86.07\big]$ \\ 
\\[-0.9em]
\textbf{300} & $\big[76.94,88.42\big]$ & $\big[84.25,93.17\big]$ & $\big[86.77,93.27\big]$ \\ 
\\[-0.9em]
\textbf{1K}  & $\big[82.78,93.97\big]$ & $\big[90.03,96.37\big]$ & $\big[91.00,96.54\big]$ \\ 
\\[-0.9em]
\textbf{3K}  & $\big[87.99,97.04\big]$ & $\big[92.71,98.06\big]$ & $\big[94.08,98.20\big]$ \\ 
\\[-0.9em]
\textbf{10K} & $\big[91.72,98.49\big]$ & $\big[94.63,98.97\big]$ & $\big[95.84,99.07\big]$ \\ 
\\[-0.9em]
\textbf{30K} & $\big[94.07,99.14\big]$ & $\big[96.38,99.38\big]$ & $\big[97.60,99.44\big]$ \\
\\[-0.9em] 
\textbf{$\approx$ 56K} & $\big[94.87,99.30\big]$ & $\big[97.39,99.51\big]$ & $\big[98.05,99.55\big]$ \\
\end{tabular}
}
\end{center}
\end{table}

\begin{table}[ht]
\begin{center}
\caption{Summary of \textit{robustness} observed on all three CNN models using varying number of training samples.}
\label{tab:dataset_robustness}
\vskip 0.1in
\resizebox{55mm}{!} {
\begin{tabular}{c|ccc}
\multirow{2}{*}{\textbf{\begin{tabular}[c]{@{}c@{}}\# of training \\samples\end{tabular}}} & \multicolumn{3}{c}{{$\rho$}}                   \\ \cline{2-4} 
\\[-1em]
& CNN-1 & CNN-2 & CNN-3 \\ \hline
\\[-0.9em]	
\textbf{100} & $0.854$ & $0.869$ & $0.882$ \\ 
\\[-1em]
\textbf{300} & $0.870$ & $0.904$ & $0.930$ \\
\\[-1em] 
\textbf{1K}  & $0.881$ & $0.934$ & $0.943$ \\ 
\\[-1em]
\textbf{3K}  & $0.907$ & $0.945$ & $0.958$ \\
\\[-1em] 
\textbf{10K} & $0.931$ & $0.956$ & $0.967$ \\
\\[-1em] 
\textbf{30K} & $0.949$ & $0.970$ & $0.981$ \\
\\[-1em] 
\textbf{$\approx$ 56K} & $0.955$ & $0.979$ & $0.985$ \\ 
\end{tabular}
}
\end{center}
\end{table}

\subsubsection{Class Label Specific Dataset Adequacy }

We have performed further analysis to understand dataset robustness on individual class label basis. More specifically, one of the classes in our dataset might have more variety among its samples and create clusters with a wider range of distributions.  In such a case, we might need to use more examples for that class to obtain a good generalization, while other classes are less variant and existing samples are enough for the algorithm to perform well even on unlikely test samples. Analyzing the class specific results of dataset robustness may help the researcher to optimize efforts for extra data collection by identifying the less robust classes.

\begin{figure}[h]
	\begin{center}
		\centerline{\includegraphics[width=\columnwidth,trim={0cm 0cm 0cm 0cm},clip]{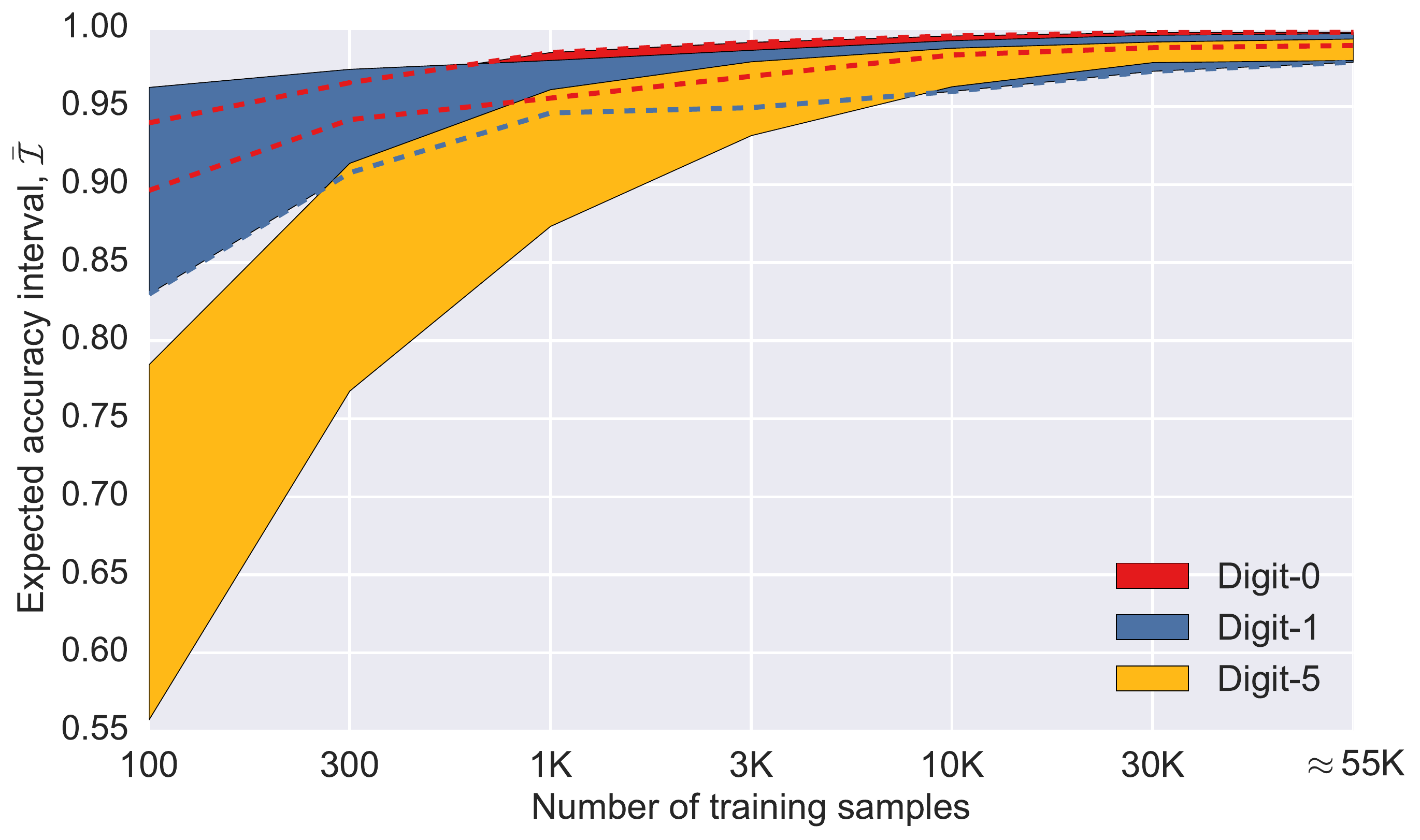}}
		\caption{Change of \textit{expected accuracy interval} on class basis with respect to the number of training samples observed on CNN-3 for digit-0, digit-1 and digit-5.}
		\label{fig:data_class_acc}
	\end{center}
	\vskip -0.3in
\end{figure}

\begin{figure}[t]
	\begin{center}
		\centerline{\includegraphics[width=\columnwidth,trim={0cm 0cm 0cm 0cm},clip]{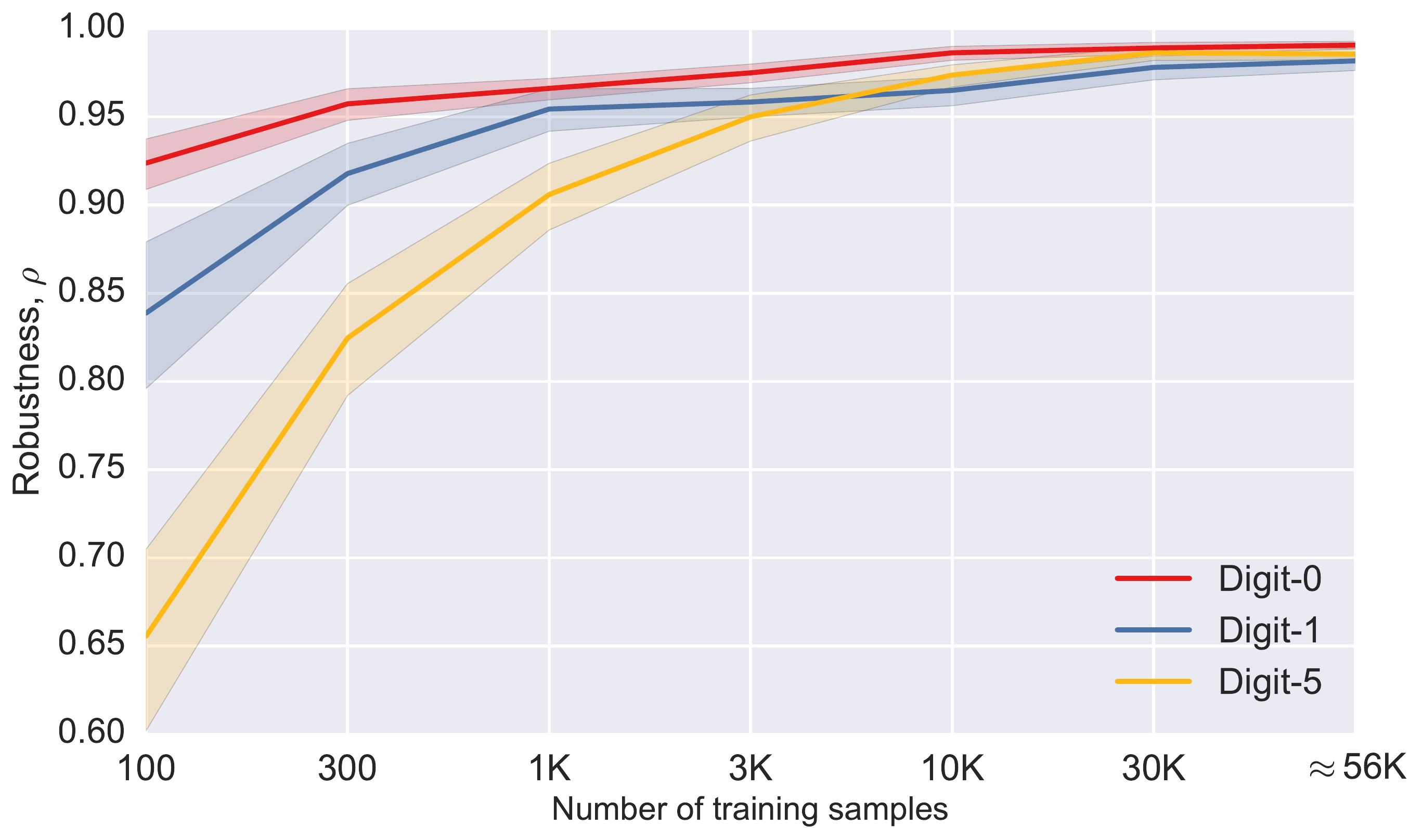}}
		\caption{Change of class specific dataset \textit{robustness} with respect to the number of training samples observed on CNN-3 for digit-0, digit-1 and digit-5.}
		\label{fig:data_class_robust}
	\end{center}
	\vskip -0.2in
\end{figure}

Figure~\ref{fig:data_class_acc} and Figure~\ref{fig:data_class_robust} illustrate the class-based \textit{expected accuracy interval} and \textit{robustness} observed on CNN-3 model for three sample classes of MNIST - digit-0, digit-1 and digit-5 - showing considerably different characteristics with respect to the increasing number of training samples. When the number of training samples is limited to 100, i.e. 10 samples/class, the expected generalization of the model is most accurate when encountering likely digit-1 test samples.  However in terms of the robustness metric, the model performs better for digit-0 where \textit{expected accuracy interval} is narrower. In other words, the model needs more examples of digit-1 in order to develop a better generalization on unlikely test samples as it has already done so for likely ones. 
On the other hand, for digit-5 class, the model does not perform well on either likely or unlikely test samples when the training set size is small. Remarkably, as more training samples are added, the model becomes more robust for digit-5 than for digit-1. Without using source-aware partitioning, one might detect the difficult class labels like digit-5. However, it is impossible to identify those behaving like digit-1 as they are likely to yield good performance on test sets constructed by random cross-validation. Source-awareness enables us to identify such classes by creating more challenging subset partitioning.    

\section{CONCLUSION}

The expected generalization error, which is traditionally calculated over a test set constructed by random subsampling from the entire dataset, can be a quick measure to compare different learning algorithms. However, it is not sufficient to make a realistic assessment of performance under real-world conditions in which encountering an unlikely input compared to those within the dataset is highly possible. To simulate such conditions, we can perform dataset partitioning considering the mutually exclusive distribution of the sources, i.e. data-generating processes, existing in the dataset. When data-sources are not readily apparent, we can approximate this kind of information by applying clustering algorithms to the samples of each class such that each cluster corresponds to a different distribution or source of its class. In this paper, we have applied two source-aware partitioning schemes to simulate two extreme cases on MNIST. We show that \textit{Inclusive source-aware} partitioning, which ensures that samples from each source are included in all subsets, is a better evaluation of the expected accuracy for the best case real-world scenarios when learning model encounters inputs that are similar to the dataset samples. On the other hand, obtained accuracy using \textit{exclusive source-aware} partitioning, which ensures the mutually exclusive use of sources, represent a better evaluation of the expected accuracy for the worst case real-world scenarios when learning model encounters inputs statistically more dissimilar to the dataset samples. We proposed using an interval defined by these two accuracy values and their ratio, \textit{expected accuracy interval} and \textit{robustness}, for more thorough real-world performance assessment. On MNIST, we have shown that the proposed evaluation can be used for two separate applications: i) more rigorous model comparison and ii) dataset adequacy evaluation and class specific dataset robustness which can help researchers optimize their efforts for data collection by identifying the less robust classes. In this paper, we have only considered two-way splitting of the dataset, i.e. training and testing. However, as future work, the proposed method can also be applied for three-way splitting, i.e. training, validation and testing. Validation set constructed by using the sources whose samples are never introduced during training can help us find better hyperparameter settings resulting in better generalization on test set or for more informed early-stopping for less overfitting on the training set. 

%\newpage

\bibliography{bibliography}
\bibliographystyle{apalike}

\end{document}